\documentclass[runningheads]{llncs}
\usepackage{graphicx}
\usepackage{comment}
\usepackage{amsmath,amssymb} 
\usepackage{color}

\usepackage{color}
\usepackage{booktabs}
\usepackage{algorithm}
\usepackage{subfig}
\usepackage{algpseudocode}
\algnewcommand{\OR}{\algorithmicor}
\algnewcommand{\AND}{\algorithmicand}
\usepackage{bm}
\usepackage{amsfonts} 
\usepackage{cases}

\newcommand{\etal}{\textit{et al. }}

\newcommand{\imbmixup}{Remix}

\begin{document}
\pagestyle{headings}
\mainmatter
\def\ECCVSubNumber{2}  


\title{Remix: Rebalanced Mixup}

\titlerunning{Rebalanced Mixup}
%
\authorrunning{Chou et al.}
%

\author{Hsin-Ping Chou$^{1}$, Shih-Chieh Chang$^{1}$, Jia-Yu Pan$^{2}$, Wei Wei$^{2}$, Da-Cheng Juan$^{2}$}

\institute{
$^1$Department of Computer Science, National Tsing-Hua University, Hsinchu, Taiwan \\ 
$^2$Google Research, Mountain View, CA, USA}

\maketitle

\begin{abstract}

Deep image classifiers often perform poorly when training data are heavily class-imbalanced. In this work, we propose a new regularization technique, \imbmixup, that relaxes Mixup's formulation and enables the mixing factors of features and labels to be disentangled. Specifically, when mixing two samples, while features are mixed in the same fashion as Mixup, \imbmixup\ assigns the label in favor of the minority class by providing a disproportionately higher weight to the minority class. By doing so, the classifier learns to push the decision boundaries towards the majority classes and balance the generalization error between majority and minority classes. We have studied the state-of-the art regularization techniques such as Mixup, Manifold Mixup and CutMix under class-imbalanced regime, and shown that the proposed \imbmixup\ significantly outperforms these state-of-the-arts and several re-weighting and re-sampling techniques, on the imbalanced datasets constructed by CIFAR-10, CIFAR-100, and CINIC-10. We have also evaluated \imbmixup\ on a real-world large-scale imbalanced dataset, iNaturalist 2018. The experimental results confirmed that \imbmixup\ provides consistent and significant improvements over previous methods. 


\end{abstract}

\section{Introduction}
Deep neural networks have made notable breakthroughs in many fields such as computer vision \cite{Wang2018VideotoVideoS,Shaham_2019_ICCV,Li2019LearningTD}, natural language processing \cite{Lample2018PhraseBasedN,Lan2019ALBERTAL,Devlin2019BERTPO} and reinforcement learning \cite{Steenkiste2019APO}. Aside from delicately-designed algorithms and architectures, training data is one of the critical factors that affects the performance of neural models. In general, training data needs to be carefully labeled and designed in a way to achieve a balanced distribution among classes. However, a common problem in practice is that certain classes may have a significantly larger presence in the training set than other classes, making the distribution skewed. Such a scenario is referred to as data imbalance. Data imbalance may bias neural networks toward the majority classes when making inferences.  

Many previous works have been proposed to mitigate this issue for training neural network models. Most of the existing works can be split into two categories: re-weighting and re-sampling. Re-weighting focuses on tuning the cost (or loss) for different classes. Re-sampling focuses on reconstructing a balanced dataset by either over-sampling the minority classes or under-sampling the majority classes. Both re-weighting and re-sampling have some disadvantages when used for deep neural networks. Re-weighting tends to make optimization difficult under extreme imbalance. Furthermore, it has been shown that re-weighting is not effective when no regularization is applied \cite{Byrd2018WhatIT}. Re-sampling is very useful in general especially for over-sampling techniques like SMOTE \cite{smote}. However, it is hard to integrate into modern deep neural networks where feature extraction and classification are performed in an end-to-end fashion while over-sampling is done subsequent to feature extraction. This issue is particularly difficult to overcome when training with large-scale datasets. 

\begin{figure}[t]
    \centering
    {\includegraphics[width=0.56\textwidth]{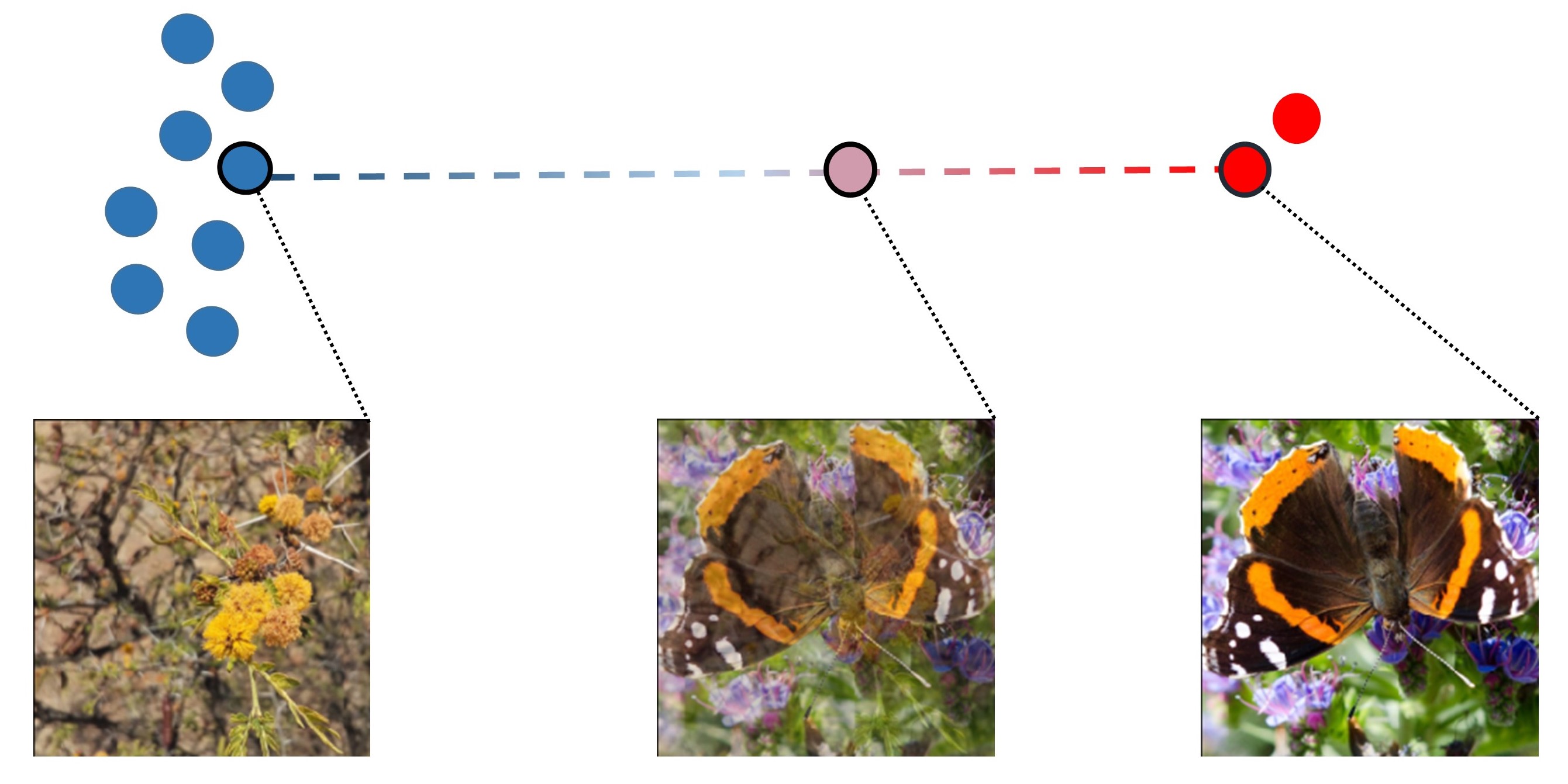}
    \caption{Assume the label ``butterfly'' belongs to the minority class and the label ``yellow plant'' belongs to the majority class. All linear combinations of the two images are on the dashed line. For a mixed image that is 70\% ``butterfly`` and 30\% ``yellow plant`` like the middle one, Mixup will assign the label to be 70\% ``butterfly`` and 30\% ``yellow plant``. However, \imbmixup\ would assign a label that is in favor of the minority class, e.g., 100\% ``butterfly``. For more details on how exactly \imbmixup\ assigns labels for the mix images, please refer to Section 3 and 4.}
    \label{fig:intro}
    }	
\end{figure}

In order to come up with a solution that is convenient to incorporate for large-scale datasets, we focus on regularization techniques which normally introduce little extra costs. Despite the recent success in regularizations \cite{cutout,mixup,manifoldmixup,cutmix}, these advanced techniques are often designed for balanced data and evaluated on commonly used datasets (e.g., CIFAR, ImageNet ILSVRC 2012) while real-world datasets tend to have a long-tailed distribution of labels \cite{Horn2017TheDI,Liu2019LargeScaleLR}. As a result, our motivation is to make the commonly used regularization techniques such as Mixup \cite{mixup}, Manifold Mixup \cite{manifoldmixup} and CutMix \cite{cutmix} perform better in the real-world imbalanced scenario. 

The key idea of Mixup is to train a neural model with mixed samples virtually created via the convex combinations of pairs of features and labels. Specifically, Mixup assumes that linear interpolations of feature vectors should come with linear interpolations of the associated labels using the same mixing factor $\lambda$. We observe that this assumption works poorly when the class distribution is imbalanced. In this work, we propose \imbmixup\ that relaxes the constraint of using the same mixing factor and allows the mixing factors of the features and labels to be different when constructing the virtual mixed samples. Fig. \ref{fig:intro} illustrates the difference between \imbmixup\ and Mixup. Note that the mixing factor of labels is selected in a way to provide a better trade-off between the majority and minority classes. 

This work brings the following contributions. (a) We propose \imbmixup, \ a computationally cheap regularization technique to improve the model generalization when training with imbalanced data. (b) The proposed \imbmixup\ can be applied to all Mixup-based regularizations and can be easily used with existing solutions against data imbalance to achieve better performance. (c) We evaluate \imbmixup\ extensively on various imbalanced settings and confirm that \imbmixup\ is general and effective for different scenarios.

\section{Related Works}
\subsubsection{Re-Weighting} Re-weighting (cost-sensitive learning) focuses on tuning cost or loss to redefine the importance of each class or sample \cite{Chung2015CostAwarePF,Wang2016TrainingDN,Huang2016LearningDR,Bul2017LossMF}. In particular, early works \cite{earlybinary1,Byrd2018WhatIT} study on how re-weighting affects the decision boundary in the binary classification case. The naive practice of dealing with an imbalanced dataset is weighted by the inverse class frequency or by the inverse square root of class frequency. Motivated by the observation that each sample might cover a small neighboring region rather than just a single point,  Cui \etal \cite{cb} introduced the concept of ``effective number'' of a class, which takes the class overlapping into consideration for re-weighting. In general, re-weighting methods perform poorly when the classes are extremely imbalanced, where the performance of majority classes is significantly compromised. As an example, Cao \etal \cite{ldam} show that re-weighting can perform even worse than vanilla training in the extreme setting.

\subsubsection{Re-Sampling}
Re-sampling methods can be summarized into two categories: over-sampling the minority classes \cite{genminority,advimb} and under-sampling the majority classes. Both of these methods have drawbacks. Over-sampling the minority classes may cause over-fitting to these samples, and under-sampling majority samples discards data and information, which is wasteful when the imbalance is extreme. For over-sampling, instead of sampling from the same group of data, augmentation techniques are applied to create synthetic samples. Classical methods include SMOTE \cite{smote} and ADASYN \cite{adasyn}. The key idea of such methods is to find the $k$ nearest neighbors of a given sample and use the interpolation to create new samples.

\subsubsection{Alternative Training Objectives}
Novel objectives are also proposed to fight against class imbalance. For example, Focal Loss \cite{focal} identifies the class imbalance in object detection task, and the authors proposed to add a modulating term to cross entropy in order to focus on hard negative examples. Although Focal Loss brought significant improvements in object detection tasks, this method is known to be less effective for large-scale imbalanced image classification \cite{ldam}. Another important work for designing an alternative objective for class imbalance is the Label-Distribution-Aware Margin Loss \cite{ldam}, which is motivated by the recent progress on various margin-based losses \cite{Liu2016LargeMarginSL,Liu2017SphereFaceDH}. Cao \etal \cite{ldam} derived a theoretical formulation to support their proposed method that encourages minority classes to have larger margins and encourage majority classes to have smaller margins. 

\subsubsection{Other Types}
Two competitive state-of-the arts \cite{bbn,decouplerepresentation} focus on the representation learning and classifier learning of a CNN. Kang \etal \cite{decouplerepresentation} found that it is possible to achieve strong long-tailed recognition ability by adjusting only the classifier. Zhou \etal \cite{bbn} proposed a Bilateral-Branch Network to take care both representation and classifier learning.

\subsubsection{Mixup-based Regularization}
Mixup \cite{mixup} is a regularization technique that proposed to train with interpolations of samples. Despite its simplicity, it works surprisingly well for improving generalization of deep neural networks. Mixup inspires several follow-up works like Manifold Mixup \cite{manifoldmixup}, RICAP \cite{RICAP} and CutMix \cite{cutmix} that have shown significant improvement over Mixup. Mixup also shed lights upon other learning tasks such as semi-supervised learning \cite{ict,mixmatch}, adversarial defense \cite{MI} and neural network calibration \cite{calimix}.

\section{Preliminaries}
\subsection{Mixup}
Mixup \cite{mixup} was proposed as a regularization technique for improving the generalization of deep neural networks. The general idea of Mixup is to generate mixed sample $\tilde{x}^{MU}$ and $\tilde{y}$ by linearly combining an arbitrary sample pair ($x_i$,$y_i$; $x_j$, $y_j$) in a dataset $\mathcal{D}$. In Eq.~\ref{mixup_x}, this mixing process is done by using a mixing factor $\lambda$ which is sampled from the beta distribution.

\begin{align}
  \tilde{x}^{MU} &= \lambda x_i + (1 - \lambda) x_j \label{mixup_x} \\  
  \tilde{y} &= \lambda y_i + (1 - \lambda) y_j \label{mixup_y}
\end{align}

\subsection{Manifold Mixup}
Instead of mixing samples in the feature space, Manifold Mixup \cite{mixup} performs the linear combination in the embedding space. This is achieved by randomly performing the linear combination at an eligible layer $k$ and conducting Mixup on $(g_k(x_i), y_i)$ and $(g_k(x_j), y_j)$ where $g_k(x_i)$ denotes a forward pass until layer $k$. As a result, the mixed representations which we denoted as $\tilde{x}^{MM}$ can be thought of "mixed samples" that is forwarded from layer $k$ to the output layer. Conducting the interpolations in deeper hidden layers which captures higher-level information provides more training signals than Mixup and thus further improve the generalization.

\begin{align}
    \tilde{x}^{MM} &= \lambda g_k(x_i) + (1 - \lambda) g_k(x_j)\label{mm_x}\\
  \tilde{y} &= \lambda y_i + (1 - \lambda) y_j
\end{align}

\subsection{CutMix}
Inspired by Mixup and Cutout \cite{cutout}, rather than mixing samples on the entire input feature space like Mixup does, CutMix \cite{cutmix} works by masking out a patch of it $B$ when generating the synthesized samples where patch $B$ is a masking box with width $r_{w} = W \sqrt{1-\lambda}$ and height $r_{h} = H \sqrt{1-\lambda}$ randomly sampled across the image. Here $W$ and $H$ are the original width and height of the image, respectively. The generated block makes sure that the proportion of the image being masked out is equal to $\frac{r_w r_h}{WH} = 1-\lambda$. A image level mask $M$ is then generated based on $B$ with elements equal to $0$ when it is blocked by $B$ and $1$ otherwise. CutMix is defined in a way similar to Mixup and Manifold Mixup in Eq.~\ref{cm_y}. Here $\odot$ is element-wise multiplication and we denote $M$ to be generated by a random process that involves $W$, $H$, and $\lambda$ using a mapping $f(\cdot)$

\begin{align}
    \Tilde{x}^{CM} & =  \mathbf{M} \odot x_{i} + (\mathbf{1}- \mathbf{M}) \odot x_{j} \label{cm_x}\\
    \Tilde{y} & =  \lambda y_i + (1-\lambda) y_j \label{cm_y}\\ 
    \mathbf{M} &\sim f(\cdot | \lambda, W, H) \label{mask}
\end{align}

\section{\imbmixup}
\label{sec:method}
We observe that both Mixup, Manifold Mixup, and CutMix use the same mixing factor $\lambda$ for mixing samples in both feature space and label space. We argue that it does not make sense under the imbalanced data regime and propose to disentangle the mixing factors. After relaxing the mixing factors, we are able to assign a higher weight to the minority class so that we can create labels that are in favor to the minority class. Before we further introduce our method. We first show the formulation of \imbmixup\ as below:

\begin{align}
  \tilde{x}^{RM} = \lambda_x x_i + (1 - \lambda_x) x_j \label{rm_x}\\
  \tilde{y}^{RM} = \lambda_y y_i + (1 - \lambda_y) y_j \label{rm_y}
\end{align}

The above formulation is in a more general form compares to other Mixup-based methods. In fact, $\tilde{x}^{RM}$ can be generated based on Mixup, Manifold Mixup, and CutMix according to Eq. \ref{mixup_x}, Eq. \ref{mm_x} and Eq. \ref{cm_x} respectively. Here we use Mixup for the above formulation as an example. Note that Eq. \ref{rm_y} relaxes the mixing factors which are otherwise tightly coupled in the original Mixup's formulation. Mixup, Manifold Mixup, and Cutmix are a special case when $\lambda_y = \lambda_x$. Again, $\lambda_x$ is sampled from the beta distribution and we define the exact form of $\lambda_y$ as in Eq.\ref{eq:lambda_y}.

\begin{equation}{\lambda_y =}
\begin{cases}
     0, &  n_i/n_j \geq \kappa \text{ and } \lambda < \tau;\  \\
     1, &  n_i/n_j \leq 1/\kappa \text{ and } 1-\lambda < \tau;\  \\
     \lambda, & \text{otherwise}   
\end{cases}
\label{eq:lambda_y}
\end{equation}

Here $n_i$ and $n_j$ denote the number of samples in the corresponding classification class from sample $i$ and sample $j$. For example, if $y_i=1$ and $y_j=10$, $n_i$ and $n_j$ would be the number of samples for class $1$ and $10$, which are the class that these two samples represent. $\kappa$ and $\tau$ are two hyper-parameters in our method. To understand what Eq.~\ref{eq:lambda_y} is about, we first define the $\kappa$-majority below. 

\begin{definition}
$\kappa$-Majority. A sample $(x_i,y_i)$,  is considered to be $\kappa$-majority than sample $(x_j,y_j)$, if $n_i/n_j \ge \kappa$ where $n_i$ and $n_j$ represent the number of samples that belong to class $y_i$ and class $y_j$, respectively.
\end{definition}

\begin{figure}[t]
\minipage{0.3\textwidth}
  \subfloat[$\tau$=0]{\includegraphics[width=\linewidth]{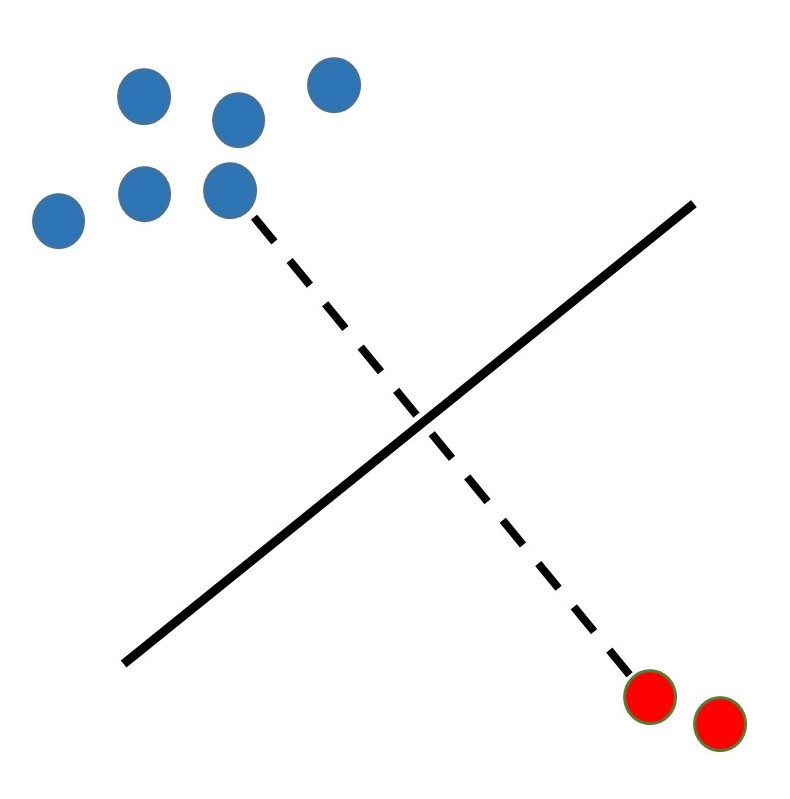}}%
\endminipage\hfill
\minipage{0.3\textwidth}
  \subfloat[ $0<\tau<1$]{\includegraphics[width=\linewidth]{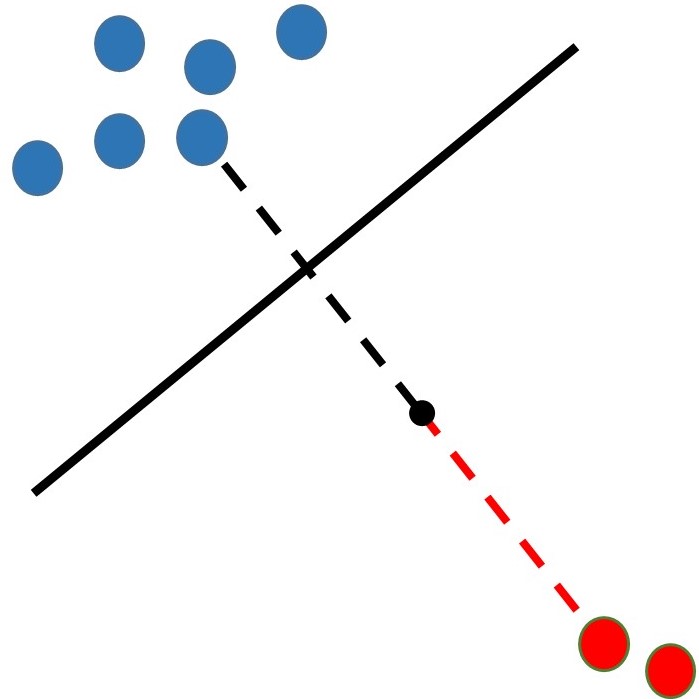}}%
\endminipage\hfill
\minipage{0.3\textwidth}%
  \subfloat[$\tau$=1]{\includegraphics[width=\linewidth]{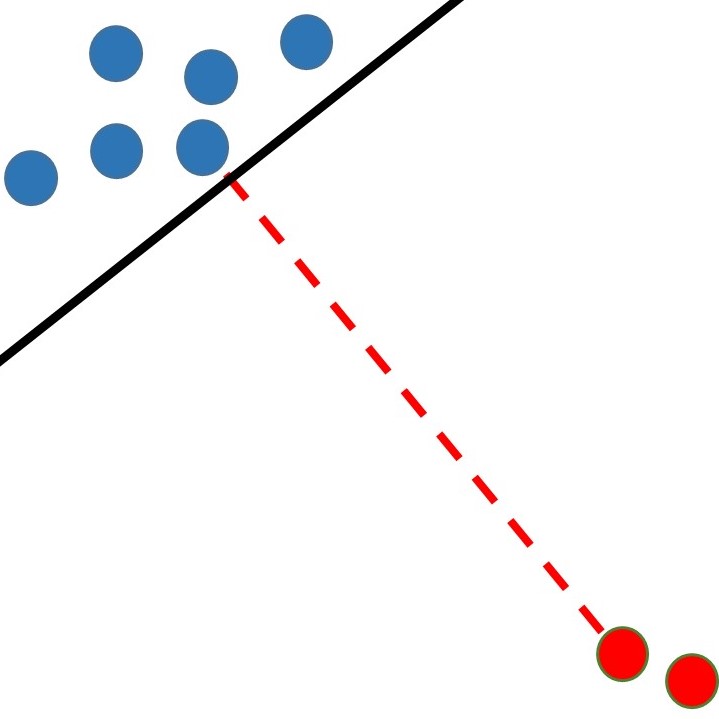}}%
\endminipage
\caption{A simple illustration of how the hyper-parameter $\tau$ affects the boundary. Blue and red dots are majority and minority samples in the feature space. The dashed line represents all the possible mixed samples and the solid black line represents the decision boundary.  When $\tau$ is set to $0$, mixed samples are linearly mixed and labelled as the original Mixup, Manifold Mixup, and Cutmix algorithms. But when $\tau$ is set to a value larger than zero, then part of the mixed samples on the red dashed line will be labelled as the minority class. In the most extreme case for fighting against data imbalance, $\tau$ is set to $1$ where all mixed samples are labelled as the minority class.}
\label{fig:proof}
\end{figure}

The general idea in Eq.~\ref{eq:lambda_y} shows when exactly \imbmixup\ assigns the synthesized labels in favor to the minority class. When $x_i$ is $\kappa$-majority to $x_j$ and the other condition is met, $\lambda_y$ is set to $0$ which makes the the synthesized labels 100\% contributed by the minority class $y_j$. Conversely, when $x_j$ is $\kappa$-majority to $x_i$ along with other conditions, $\lambda_y$ to $1$ which makes the the synthesized labels 100\% contributed by the minority class $y_i$. 

The reason behind this choice of making the synthesized samples to be labeled as the minority class is to move the decision boundary towards the majority class. This is aligned with the consensus in the community of imbalanced classification problem. In \cite{ldam}, the authors gave a rather theoretical analysis illustrating that how exactly the decision boundary should be pushed towards the majority classes by using a margin loss. Because pushing the decision boundary towards too much may hurt the performance of the majority class. As a result, we don't want the synthesized labels to be always pointing to the minority class whenever mixing a majority class sample and a minority class sample. To achieve that we have introduced another condition in Eq.~\ref{eq:lambda_y} controlled by parameter $\tau$ that is conditioned on $\lambda$. In both conditions, extreme cases will be rejected and $\lambda_y$ will be set to $\lambda$ when $\lambda_x$ is smaller than $\tau$. The geometric interpretation of this condition can be visualized in Fig.~\ref{fig:proof}. Here we see that when $\tau$ is set to $0$, our approach will degenerate to the base method, which can be Mixup, Manifold Mixup, or Cutmix depending on the choice of design. When $\tau$ is set to a value that is larger than $0$, synthesized samples $\tilde{x}^{RM}$ that are close to the minority classes will be labelled as the minority class, thus benefiting the imbalanced classification problems. To summarize, $\tau$ controls the extent that the minority samples would dominate the label of the synthesized samples. When the conditions are not met, or in other words, when there is no trade-off to be made, we will just use the base method to generate the labels. This can be illustrated in the last condition in Eq.~\ref{eq:lambda_y}, when none of $i$ or $j$ can claim to be $\kappa$-majority over each other and in this case $\lambda_y$ is set to $\lambda_x$.

Attentive readers may realize that using the same $\tau$ for various pairs of a majority class and a minority class implies that we want to enforce the same trade-off for those pairs. One may wonder why not introduce $\tau_{ij}$ for each pair of classes? This is because the trade-off for multi-class problem is intractable to find. Hence, instead of defining $\tau_{ij}$ for each pair of classes, we use a single $\tau$ for all pairs of classes. Despite its simplicity, using a single $\tau$ is sufficient to achieve a better trade-off.

\imbmixup\ might look similar to SMOTE \cite{smote} and ADASYN \cite{adasyn} at first glance, but they are very different in two perspectives. First, the interpolation of \imbmixup\ can be conducted with any two given samples while SMOTE and ADASYN rely on the knowledge of a sample's same-class neighbors before conducting the interpolation. Moreover, rather than only focusing on creating new data points, \imbmixup\ also pays attention to labelling the mixed data which is not an issue to SMOTE since the interpolation is conducted between same-class data points. Secondly, \imbmixup\ follows Mixup to train the classifier only on the mixed samples while SMOTE and ADASYN train the classifier on both original data and synthetic data.

To give a straightforward explanation of why \imbmixup\ would benefit learning with imbalance datasets, consider a mixed example $\Tilde{x}$ between a majority class and a minority class, the mixed sample includes features of both classes yet we mark it as the minority class more. This force the neural network model to learn that when there are features of a majority class and a minority class appearing in the sample, it should more likely to consider it as the minority class. This means that the classifier is being less strict to the minority class. Please see Section 4.7 for qualitative analysis. Note that \imbmixup\ method is a relaxation technique, and thus may be integrated with other techniques.  In the following experiments, besides showing the results of the pure \imbmixup\ method, we also show that the \imbmixup\ method can work together with the re-weighting or the re-sampling techniques.
Algorithm \ref{alg:final} shows the pseudo-code of the proposed \imbmixup\ method.

\begin{algorithm}[t]
\caption{\imbmixup} \label{alg:final}
\begin{algorithmic}[1]

\Require Dataset $\mathcal{D} = \{(x_i,y_i)\}_{i=1}^n$. A model with parameter $\theta$
\State Initialize the model parameters $\theta$ randomly
\While {$\theta$ is not converged}
	\State  $\{(x_i, y_i), (x_j, y_j)\}_{m=1}^M\leftarrow \text{SamplePairs}(\mathcal{D}, M)$ 
    \State $\lambda_x$ $\sim$ $Beta(\alpha, \alpha)$
    \For {$m=1$ to $M$}
    
        \State $\tilde{x}^{RM}$  $\leftarrow$ RemixImage($x_i, x_j, \lambda_x$) according to Eq.\ref{rm_x}
         \State $\lambda_y$ $\leftarrow$ LabelMixingFactor($\lambda_x, n_i, n_k, \tau, \kappa$) according to Eq.\ref{eq:lambda_y}
        \State $\tilde{y}^{RM}$  $\leftarrow$ RemixLabel($y_i, y_j, \lambda_y$) according to Eq.\ref{rm_y}

    \EndFor
    \State $\mathcal{L}(\theta) \leftarrow \frac{1}{M} \sum_{(\tilde{x},\tilde{y})}L((\tilde{x},\tilde{y});\theta) $ 
    \State $\theta \leftarrow \theta - \delta\nabla_\theta \mathcal{L}(\theta)$ 

\EndWhile

\end{algorithmic}
\end{algorithm}

\section{Experiments}
\begin{figure}[t]
\minipage{0.33\textwidth}
  \subfloat[Long-tail: $\rho=100$]{\includegraphics[width=\linewidth]{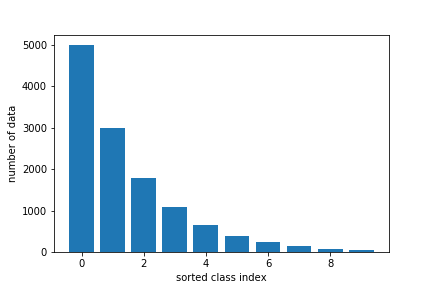}}%
\endminipage\hfill
\minipage{0.33\textwidth}
  \subfloat[Step: $\rho=10$, $ \mu=0.5$]{\includegraphics[width=\linewidth]{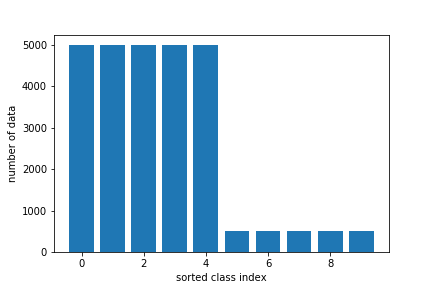}}%
\endminipage\hfill
\minipage{0.33\textwidth}%
  \subfloat[iNaturalist 2018]{\includegraphics[width=\linewidth]{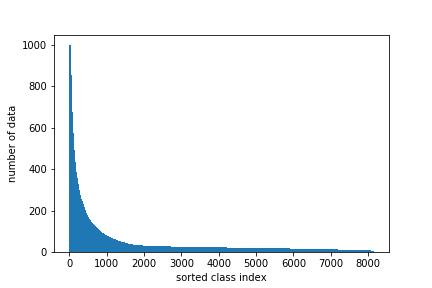}}%
\endminipage
\caption{Histograms of three imbalanced class distributions. (a) and (b) are synthesized from long-tail and step imbalance, respectively. (c) represents the class distribution of iNaturalist 2018 dataset.}
\label{fig:classimb}
\end{figure}

\subsection{Datasets}
We compare the proposed \imbmixup\ with state-of-the-art methods fighting against class imbalance on the following datasets: (a)  artificially created imbalanced datasets using CIFAR-10, CIFAR-100 and CINIC-10 datasets, and (b) iNaturalist 2018, a real-world and large-scale imbalanced dataset. 

\subsubsection{Imbalanced CIFAR}
The original CIFAR-10 and CIFAR-100 datasets both contain 50,000 training images and 10,000 validation images of size $32\times32$, with 10 and 100 classes, respectively. We follow \cite{ldam} to construct class-imbalanced datasets from the CIFAR-10 and CIFAR-100 datasets with two common imbalance types ``long-tailed imbalance" and  ``step imbalance". The validation set is kept balanced as original. Fig. \ref{fig:classimb}(a)(b) illustrates the two imbalance types. For a dataset with long-tailed imbalance, the class sizes (number of samples in the class) of the dataset follow an exponential decay. For a dataset with step imbalance, a parameter $\mu$ is used to denotes the fraction of minority classes. $\mu$ is set to 0.5 \cite{ldam} for all of the experiments. The parameter $\rho$ of the constructed datasets denotes the imbalance ratio between the number of samples from the most frequent and the least frequent classes, i.e., $\rho = \max_i \{n_i\} / \min_j \{n_j\}$.

\subsubsection{Imbalanced CINIC}
The CINIC-10 dataset \cite{cinic} is compiled by combining CIFAR-10 images with images downsampled from the ImageNet database. It contains 270,000 images and it splits train, validation, and test set equally. We only use the official split of training and validation. Using the CINIC-10 dataset helps us compare different methods better because it has 9000 training data per class which allows us to conduct extensive experiments with various imbalance ratios while making sure each class still preserves a certain number of data. This helps us focus more on the imbalance between classes rather than solving a few-shot classification problem for the minority classes. 

\subsubsection{iNaturalist 2018}
The iNaturalist species classification dataset \cite{inat} is a real-world large-scale imbalanced dataset which has 437,513 training images of 8,142 classes.  The dataset features many visually similar species which are extremely difficult to accurately classify without expert knowledge. We adopt the official training and validation splits for our experiments where the training datasets have a long-tailed label distribution and the validation set is designed to have a balanced distribution. 


\subsubsection{CIFAR and CINIC-10}
For fair comparisons, we ported the official code of \cite{ldam} into our codebase. We follow \cite{cb,ldam} use ResNet-32 for all CIFAR experiments and we use ResNet-18 for all CINIC experiments.  We train 300 epochs and decay the learning rate 0.01 at 150, 225 epoch. We use stochastic gradient descent (SGD) with momentum 0.9 and weight decay 0.0002. For non-LDAM methods, We train it for 300 epochs with mini-batch size 128. We decay our learning rate by 0.1 at 150, 250 epoch. All CIFAR and CINIC experiment results are mean over 5 runs. Standard data augmentation is applied, which is the combination of random crop, random horizontal flip and normalization. If DRW or DRS are used for the training, we use re-weighting or re-sampling at the second learning rate decay.  We set $\tau=0.5$ and $\kappa$ = 3 for all experiments. The choice of these two parameters can be found with simple grid search. Despite that one might want to use carefully-tuned 
parameters for different imbalance scenarios, we empirically found that setting $\tau=0.5$ and $\kappa$ = 3 is able to provide consistent improvements over the previous state-of-the-arts.

\subsubsection{iNaturalist 2018} 
We use ResNet-50 as the backbone network across all experiments for iNaturalist 2018. Each image is first resized to 
$256\times256$, and then a $224\times224$ crop is randomly sampled from an image or its horizontal flip. Then color jittering and lighting are applied. Follow \cite{cb,ldam}, we train the network for 90 epochs with an initial learning rate of 0.1 and mini-batch size 256. We anneal the learning rate at epoch 30 and 60. For the longer training schedule, we train the network for 200 epochs with an initial learning rate of 0.1 and anneal the learning rate at epoch 75 and 150. Using the longer training schedule is necessary for Mixup-based regularizations to converge \cite{mixup,cutmix}. We set $\tau=0.5$ and $\kappa$ = 3.

\begin{table}[t]
	\centering
	\caption{Top-1 accuracy of on imbalanced CIFAR-10 and
	CIFAR-100. \\        
	$\dagger$ denotes the results from the original paper.}
	\label{tab:imbtable2}
	\begin{tabular}{c|cc|cc|cc|cc}
		\toprule
		Dataset           & \multicolumn{4}{c|}{Imbalanced CIFAR-10}                               & \multicolumn{4}{c}{Imbalanced CIFAR-100}                              \\ \midrule
		Imbalance Type         & \multicolumn{2}{c|}{long-tailed} & \multicolumn{2}{c|}{step} & \multicolumn{2}{c|}{long-tailed} & \multicolumn{2}{c}{step} \\ \midrule
		Imbalance Ratio        & \multicolumn{1}{c|}{100}  & 10 & \multicolumn{1}{c|}{100}   & 10   & \multicolumn{1}{c|}{100}  & 10 & \multicolumn{1}{c|}{100}   & 10   \\ \midrule
		ERM & 71.86 & 86.22 & 64.17 & 84.02 & 40.12 & 56.77 &  40.13 & 54.74 \\
		Focal \cite{focal} $\dagger$ & 70.18 & 86.66 & 63.91 & 83.64 &38.41 &55.78 &38.57 & 53.27\\
	    RW Focal \cite{cb} $\dagger$ &74.57 &87.1 &60.27 &83.46 &36.02 &57.99 &19.75 &50.02\\
	    DRS \cite{ldam} $\dagger$ & 74.50 & 86.72  & 72.03 & 85.17 & 40.33 & 57.26 & 41.35 & 56.79\\
	    DRW \cite{ldam} $\dagger$ & 74.86 & 86.88 & 71.60 & 85.51 & 40.66 & 57.32 & 41.14 & 57.22\\
	    LDAM \cite{ldam} $\dagger$ & 73.35 & 86.96 & 66.58 & 85.00 & 39.60 & 56.91 & 39.58 & 56.27  \\
		LDAM-DRW \cite{ldam}  & 76.57 & 86.7 & 75.94 & 86.52 & 42.64 & 57.18 & 45.40 & 57.09 \\
		Mixup \cite{mixup} & 73.09 & 88.00 & 65.80 & 85.20 & 40.83 & 58.37 &  39.64 & 54.46 \\
		\imbmixup & 75.36 & 88.15 & 68.98 & 86.34 & 41.94 & 59.36 & 39.96 & 57.06 \\
		BBN \cite{bbn} $\dagger$ & \textbf{79.82} & 88.32 & -- & -- & 42.56 & 59.12 & -- & -- \\
		
		\midrule
		\imbmixup-LDAM-DRW & 79.33 & 86.78 & 77.81 & 86.46  & 45.02 & 
		59.47 & 45.32 & 56.59 \\
		\imbmixup-LDAM-DRS & 79.45 & 87.16 & 78.00 & 86.91  & 45.66 & 
		59.21 & 45.74 & 56.19 \\
		\midrule
		\imbmixup-RS & 76.23 & 87.70 & 67.28 & 86.63  & 41.13 & 
		58.62 & 39.74  & 56.09 \\
        \imbmixup-RW & 75.1 & 87.91 & 68.74 & 86.38  & 33.51  & 57.65
		& 17.42 & 54.45  \\
		\midrule
		\imbmixup-DRS & 79.53 & 88.85 & 77.46 & 88.16 & 46.53 & 
		60.52 & \textbf{47.25} & \textbf{60.76} \\
		\imbmixup-DRW & 79.76 & \textbf{89.02} & \textbf{77.86} & \textbf{88.34}  & \textbf{46.77} & \textbf{61.23}
		& 46.78 & 60.44   \\
		\bottomrule
	\end{tabular}
\end{table}

\subsubsection{Baseline Methods for Comparison}
We compare our methods with vanilla training, state-of-the-art techniques and their combinations. 
(1) Empirical risk minimization (ERM): Standard training with no anti-imbalance techniques involved. (2) Focal: Use focal loss instead of cross entropy. (3) Re-weighting (RW): Re-weight each sample by the effective number which is defined as $E_n = (1-\beta^n)/(1-\beta)$, where $\beta = (N-1)/N$. (4) Re-sampling (RS): Each example is sampled with probability proportional to the inverse of effective number. (5) Deferred re-weighting and deferred re-sampling (DRW, DRS): A deferred training procedure which first trains using ERM before annealing the learning rate, and then deploys RW or RS (6) LDAM: A label-distribution-aware margin loss which considers the trade-off between the class margins. (7) Mixup: Each batch of training samples are generated according to \cite{mixup} (8) BBN \cite{bbn} and LWS \cite{decouplerepresentation}: Two state-of-the-arts methods. We directly copy the results from the original paper.

\begin{table}[t]
\centering
\caption{Top-1 accuracy on imbalanced CINIC-10 using ResNet-18.}
\begin{tabular}{@{}c|cccc|cccc@{}}
\toprule
Imbalance Type & \multicolumn{4}{c|}{long-tailed} & \multicolumn{4}{c}{step} \\ \midrule
Imbalance Ratio & \multicolumn{1}{c|}{200} & \multicolumn{1}{c|}{100} & \multicolumn{1}{c|}{50} & 10 & \multicolumn{1}{c|}{200} & \multicolumn{1}{c|}{100} & \multicolumn{1}{c|}{50} & 10 \\ \midrule
ERM & 56.16 & 61.82 & 72.34 & 77.06 & 51.64 & 55.64 & 68.35 & 74.16 \\
RS \cite{cb} & 53.71 & 59.11 & 71.28 & 75.99 & 50.65 & 53.82 & 65.54 & 71.33 \\
RW \cite{cb} & 54.84 & 60.87 & 72.62 & 76.88 & 50.47 & 55.91 & 69.24 & 74.81 \\
DRW \cite{ldam} & 59.66 & 63.14 & 73.56 & 77.88 & 54.41 & 57.87 & 68.76 & 72.85 \\
DRS \cite{ldam} & 57.98 & 62.16 & 73.14 & 77.39 & 52.67 & 57.41 & 69.52 & 75.89 \\
Mixup \cite{mixup} & 57.93 & 62.06 & 74.55 & 79.28 & 53.47 & 56.91 & 69.74 & 75.59  \\
\imbmixup & 58.86 & 63.21 & 75.07 & 79.02 & 54.22 & 57.57 & 70.21 & 76.37 \\
LDAM-DRW \cite{ldam}& 60.80 & 65.51 & 74.94 & 77.90 & 54.93 & 61.17 & 72.26 & 76.12  \\
\midrule
\imbmixup-DRS & 61.64 & 65.95 & 75.34 & 79.17 & 60.12 & 66.53 & 75.47 & \textbf{79.86} \\
\imbmixup-DRW  & \textbf{62.95} & \textbf{67.76} & \textbf{75.49} & \textbf{79.43} & \textbf{62.82} & \textbf{67.56} & \textbf{76.55} & 79.36\\
\bottomrule
\end{tabular}

\label{tab:cinic-table}
\end{table}

\subsection{Results on Imbalanced CIFAR and CINIC}
In Table \ref{tab:imbtable2} and Table \ref{tab:cinic-table}, we compare the previous state-of-the-arts with the pure \imbmixup\ method and a variety of the \imbmixup-integrated methods. Specifically, we first integrate \imbmixup\ with basic re-weighting and re-sampling techniques with respect to the effective number \cite{cb}, and we use the deferred version of them \cite{ldam}. We also experiment with a variety that integrate our method with the LDAM loss \cite{ldam}. We observe that \imbmixup\ works particularly well with re-weighting and re-sampling. Among all the methods that we experiment with, the best performance was achieved by the method that integrates \imbmixup\ with either the deferred re-sampling method (\imbmixup-DRS) or the deferred re-weighting method (\imbmixup-DRW).

Regarding the reason why \imbmixup-DRS and \imbmixup-DRW achieve the best performance, we provide an intuitive explanation as the following: We believe that the improvement of our \imbmixup\ method comes from the imbalance-aware mixing equation for labels (Eq.\ref{eq:lambda_y}), particularly when $\lambda_y$ is set to either 0 or 1.  The more often the conditions are satisfied to make $\lambda_y$ be either 0 or 1, the more opportunity is given to the learning algorithm to adjust for the data imbalance.  To increase the chance that those conditions are satisfied, we need to have more pairs of training samples where each pair is consisted of one sample of a majority class and one sample of a minority class.  Since, by the definition of a minority class, there are not many samples from a minority class, the chance of forming such pairs of training data may not be very high.

When the re-sampling method is used with \imbmixup, the re-sampling method increases the probability of having data of minority classes in the training batches.  With more samples from minority classes in the training batches, the chance of forming sample pairs that satisfied the conditions is increased, thus allows \imbmixup\ to provide better trade-off on the data imbalance among classes.

Likewise, although using \imbmixup\ with re-weighting does not directly increase the probability of pairs that satisfy the conditions, the weights assigned to the minority classes will amplify the effect when a case of majority-minority sample pair is encountered, and thus, will also guide the classifier to have better trade-off on the data imbalance. 

On the other hand, using the LDAM loss doesn't further improve the performance. We suspect that the trade-off LDAM intends to make is competing with the trade-off \imbmixup\ intends to make. Therefore, for the rest of the experiments, we will focus more on the methods where DRS and DRW are integrated with \imbmixup.

\subsection{Results on iNaturalist 2018}
In Table \ref{tab:inat-table}, we present the results in similar order as Table \ref{tab:cinic-table}. The results show the same trends to the results on CIFAR and CINIC. Again, our proposed method, \imbmixup,\ outperforms the original Mixup. Also, state-of-the-art results are achieved when \imbmixup\ is used with re-weighting or re-sampling techniques. The model's performance is significantly better than the previous state-of-the-arts and outperforms the baseline (ERM) by a large margin. The improvement is significant and more importantly, the training cost remains almost the same in terms of training time.

\begin{table}[t]
\parbox{.45\linewidth}{
\centering

\caption{Validation errors on iNaturalist 2018 using ResNet-50.}
\begin{tabular}{cc|cc}
	\toprule
	Loss & Schedule & Top-1 & Top-5 \\
	\midrule
    ERM & SGD        & 40.19 & 18.99 \\
	RW Focal & SGD   & 38.88 & 18.97 \\
	Mixup & SGD       & 39.69 & 17.88 \\
	\imbmixup & SGD       & 38.69 & 17.70 \\
    LDAM & DRW       & 32.89 & 15.20\\
    BBN \cite{bbn} $\dagger$& --- &30.38 & ---\\
    \midrule
    LWS (200 epoch) \cite{decouplerepresentation} $\dagger$& --- &30.5 & ---\\
    LDAM (200 epoch) & DRW & 31.42 & 14.68 \\
	\imbmixup\ (200 epoch) & DRS  & \textbf{29.26}  & \textbf{12.55} \\
	\imbmixup\ (200 epoch) & DRW  & \textbf{29.51}  & \textbf{12.73} \\
    \bottomrule
    \label{tab:inat-table}
\end{tabular}

}
\hfill
\parbox{.45\linewidth}{
\centering
\caption{Top-1 accuracy of ResNet-18 trained with imbalanced CIFAR-10 with imbalance ratio $\rho$=100}
\begin{tabular}{@{}ccc@{}}
\toprule
\multicolumn{1}{c|}{Imbalance Type} & \multicolumn{1}{c|}{long-tailed} & step \\ \midrule
\multicolumn{1}{c|}{Mixup-DRW}    & \multicolumn{1}{c|}{81.09}            & 76.13  \\
\multicolumn{1}{c|}{Remix-DRW}    & \multicolumn{1}{c|}{\textbf{81.60}}            & \textbf{79.35} \\
\midrule
\multicolumn{1}{c|}{Mixup-DRS}    & \multicolumn{1}{c|}{80.40}            & 75.58     \\
\multicolumn{1}{c|}{Remix-DRS}    & \multicolumn{1}{c|}{\textbf{81.11}}     & \textbf{79.35} \\

\bottomrule
\label{tab:ablation2}
\end{tabular}

}
\end{table}

\subsection{Ablation Studies}
When it comes to addressing the issue of data imbalance, one common approach is to simply combine the re-sampling or re-weighting method with Mixup. In Table \ref{tab:ablation2}, we show the comparison of Mixup and \imbmixup\ when they are integrated with the re-sampling technique (DRS) or the re-weighting technique (DRW).  Note that our methods still outperform Mixup-based methods. The results in Table \ref{tab:ablation2} imply that, while \imbmixup\ can be considered as over-sampling the minority classes in the label space, the performance gain from \imbmixup\ does not completely overlap with the gains from the re-weighting or re-sampling techniques. 

We also observe that the improvement is more significant on datasets with step imbalance, than it is on datasets with long-tailed imbalance. In the long-tailed setting, the distribution of the class sizes makes it less likely to have pairs of data samples that satisfy the conditions of Eq.\ref{eq:lambda_y} to make $\lambda_y$ either $0$ or $1$.  On the other hand, the conditions of Eq.\ref{eq:lambda_y} are relatively more likely to be satisfied, on a dataset with the step imbalance. In other words, there is less room for \imbmixup\ to unleash its power on a dataset with long-tailed setting. 

The proposed \imbmixup\ method is general and can be applied with other  Mixup-based regularizations, such as Manifold Mixup and CutMix. In Table \ref{tab:imbtable}, we show that, the performance of Manifold Mixup and CutMix increases when they employ the \imbmixup\ regularization, significantly outperforming the performance of the vanilla Manifold Mixup or CutMix. 

Moreover, we also observe that when the imbalance ratio is not very extreme ($\rho$=10), using Mixup or \imbmixup\ doesn't produce much difference. However, when the imbalance is extreme (e.g., $\rho$=100), employing our proposed method is significantly better than the vanilla version of Manifold Mixup or CutMix. 

\begin{figure}[t]
\centering
{\includegraphics[width=\textwidth]{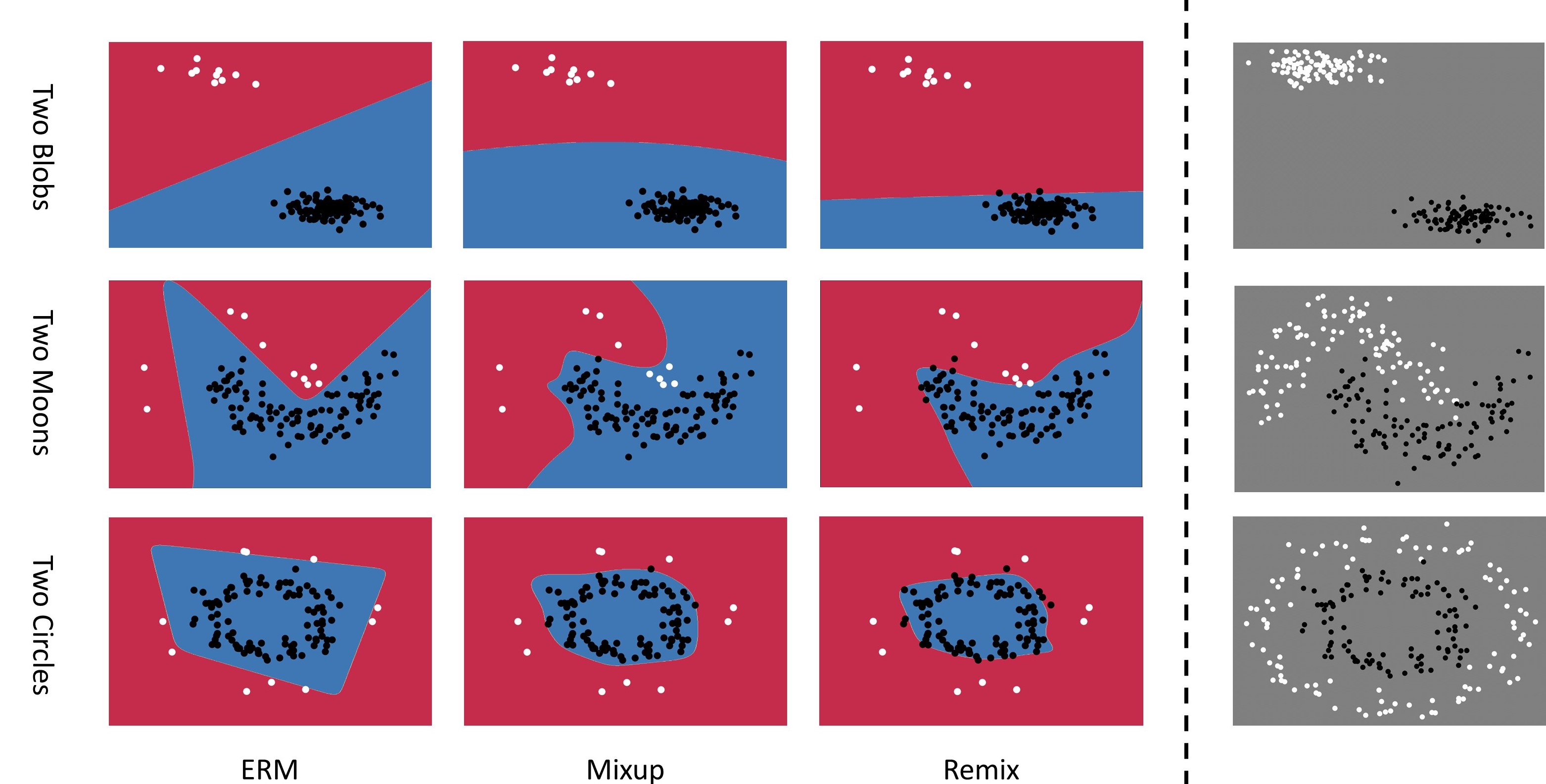}}
\caption{
A visualization of the decision boundary of the models learned by different methods. \imbmixup\ creates tighter margin for the majority class, and compensates the effect from the data imbalance.
}
\label{fig:decisionboundary}
\end{figure}

\begin{table}
	\caption{Top-1 accuracy of ResNet-32 on imbalanced CIFAR-10 and CIFAR-100.}
	\label{tab:imbtable}
	\centering
	\begin{tabular}{c|cc|cc|cc|cc}
		\toprule
		Dataset           & \multicolumn{4}{c|}{Imbalanced CIFAR-10}                               & \multicolumn{4}{c}{Imbalanced CIFAR-100}                              \\ \midrule
		Imbalance Type         & \multicolumn{2}{c|}{long-tailed} & \multicolumn{2}{c|}{step} & \multicolumn{2}{c|}{long-tailed} & \multicolumn{2}{c}{step} \\ \midrule
		Imbalance Ratio        & \multicolumn{1}{c|}{100}  & 10 & \multicolumn{1}{c|}{100}   & 10   & \multicolumn{1}{c|}{100}  & 10 & \multicolumn{1}{c|}{100}   & 10   \\ \midrule
		ERM & 71.86 & 86.22 & 64.17 & 84.02 & 40.12 & 56.77 &  40.13 & 54.74 \\
		\midrule
		Mixup \cite{mixup} & 73.09 & 88.00 & 65.80 & 85.20 & 40.83 & 58.37 &  39.64 & 54.46 \\
		\imbmixup & \textbf{75.36} & \textbf{88.15} & \textbf{68.98} & \textbf{86.34}  & \textbf{41.94} & \textbf{59.36} & \textbf{39.96} & \textbf{57.06} \\
	    \midrule
	    Manifold Mixup \cite{manifoldmixup} & 73.47 & 87.78 & 66.13 & 85.22 & 41.19 & 58.55 & 39.52 & 53.72 \\
		\imbmixup-MM & \textbf{77.07} & \textbf{88.70} & \textbf{69.78} & \textbf{87.39}  & \textbf{44.12} & \textbf{60.76} & \textbf{40.22} & \textbf{58.01} \\
		\midrule
		CutMix \cite{cutmix} & 75.04 & 88.30 & 67.97 & 86.35 & 41.86 & 59.47 & 40.23 & 56.59 \\
		\imbmixup-CM & \textbf{76.59} & 87.96 & \textbf{69.61} & \textbf{87.59}  & \textbf{43.55} & \textbf{60.15} & \textbf{40.336} &\textbf{57.78}\\

		\bottomrule
	\end{tabular}
\end{table}

\subsection{Qualitative Analysis}

To further demonstrate the effect of \imbmixup\, we present the results of \imbmixup\ on the commonly used datasets, namely, ``two blobs", ``two moons" , and ``two circles" datasets from scikit-learn \cite{scikit-learn}. In Fig.\ref{fig:decisionboundary}, the original balanced datasets are shown at the rightmost column. The three columns at the left show the created imbalanced datasets with the imbalance ratio $\rho$=10. The majority class is plotted in black and the minority class is plotted in white. The results in Fig.\ref{fig:decisionboundary} show that \imbmixup\ creates tighter margin for the majority class. In all three cases, we observe that even though our regularization sacrifices some training accuracy for the majority class (some black dots are misclassified), but it actually provides a better decision boundary for the minority class. 


\section{Conclusions and Future Work}
In this paper, we redesigned the Mixup-based regularizations for imbalanced data, called \imbmixup .\ It relaxes the mixing factor which results in pushing the decision boundaries towards majority classes. Our method is easy to implement, end-to-end trainable and computation efficient which are critical for training on large-scale imbalanced datasets. We also show that it can be easily used with existing techniques to achieve superior performance. Despite the ease of use and effectiveness, the current analysis is rather intuitive. Our future work is to dig into the mechanism and hopefully provide more in-depth analysis and theoretical guarantees which may shed lights on a more ideal form of Remix as the current one involves two hyper-parameters.


\section*{Supplementary Materials}
\subsection*{Questions for Using Remix}
In this section, we highlight things that one may wonder when using Remix. We have some observations on these questions but did not explore in sufficient depth. 

\subsubsection{How should $\tau$ and $\kappa$ be chosen?}
Simple grid search is used to find proper values for them. We first fix $\tau$ to 0 and search $\kappa$. We have done this for both step imbalance and long-tail imbalance. It is rather easy to tune $\kappa$ for step imbalance since we are dealing with "binary" imbalance. As long as the value can differentiate the minority from majority, Remix would function as we expect. After finding the value for $\kappa$, we then fix kappa and search for $\tau$. Note that we have used the same hyper-parameter setting for most experiments because we want to show that Remix is not very sensitive to the hyper-parameters. Below is a very simple ablation studies on $\tau$. It is evaluated on CIFAR-10 with $\rho$=100 with Remix-DRW. To our surprise, the hyper-parameter $\tau$ isn't as sensitive as we expected. However, based on our studies on toy dataset two blobs two moon and two circles with a MLP, we found that $\tau$ is sensitive for the exact position of the decision boundary. As a results, we think that the reason it is not that sensitive when used with ResNets is because ResNets are highly non-linear. The 

\begin{table}[]
\centering
\begin{tabular}{@{}c|c|c|c|c|c|c|c|c|c|cl@{}}
\cmidrule(r){1-11}
$\tau$      & 0.0  & 0.1  & 0.2  & 0.3  & 0.4  & 0.5  & 0.6  & 0.7  & 0.8  & 0.9  &  \\ \cmidrule(r){1-11}
Accuracy & 76.1 & 76.4 & 76.4 & 77.2 & 77.4 & 77.3 & 77.5 & 77.8 & 77.5 & 77.6 &  \\ \cmidrule(r){1-11}
\end{tabular}
\end{table}

\subsubsection{How does the original hyper-parameter $\alpha$ from Mixup affects the performance of Remix?}
Remix is greatly based on Mixup. As a result, in order to remix to work, one has to first make sure that Mixup is able to improve the performance using a certain value for $\alpha$. And since we use commonly used datasets which Mixup was also evaluated on, we were able to use the same $\alpha$ as them, which are 1.0 for CIFAR, CINIC and 0.4 for iNaturalist 2018 (Since iNaturalist 2018 images are in the same size as ImageNet, we use the same value)

\subsubsection{Does the amount of times that Remix falls into the first and second condition in eq.10 matter?}

Since the third condition makes Remix degenerate to Mixup, one may wonder if the number of times that it doesn't degenerate matter. Our intuition is that since Remix works fairly well with re-weighting and re-sampling, it matters. For example, when Remix is used with re-sampling, the chance that a majority sample and minority sample are used to create a mixed sample is greatly increased and thus increase the number of times that Remix falls into the first and second condition in eq.10. 

\begin{figure}[t]
\minipage{0.45\textwidth}
  \subfloat[ERM]{\includegraphics[width=\linewidth]{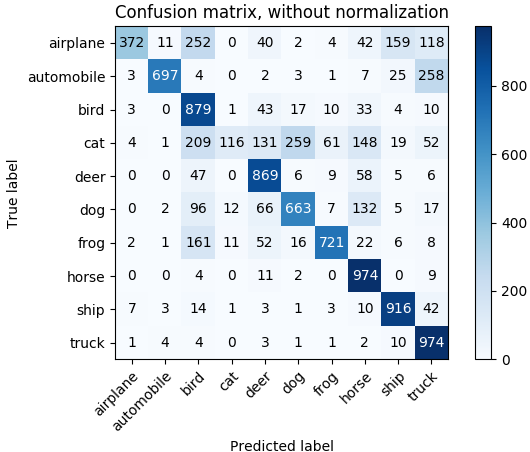}}%
\endminipage\hfill
\minipage{0.45\textwidth}%
  \subfloat[Remix]{\includegraphics[width=\linewidth]{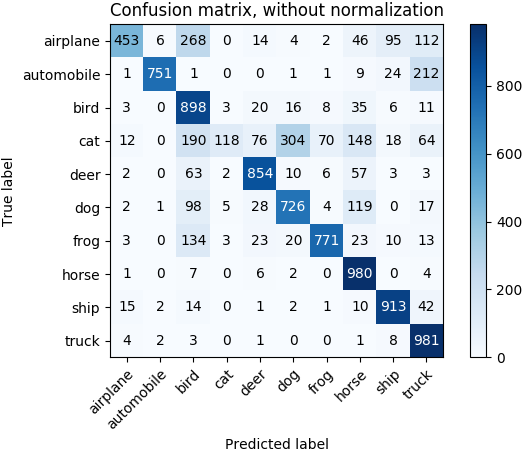}}%
\endminipage
\caption{To show that the improvement in overall accuracy is coming from the improvement on minority classes, we show the confusion matrices on CIFAR-10.}
\label{fig:proof}
\end{figure}


%
%
\bibliographystyle{splncs04}
\bibliography{egbib}
\end{document}